\documentclass[twoside,11pt]{article}

\usepackage{jmlr2e}

\usepackage{cleveref}
\usepackage{acronym}
\usepackage{enumitem}
\usepackage{todonotes}
\usepackage{pgffor}
\usepackage{xstring}
\usepackage{fancybox}
\usepackage{breakurl}
\usepackage{listings}
\usepackage{xcolor}
 
\definecolor{codegreen}{rgb}{0,0.6,0}
\definecolor{codegray}{rgb}{0.5,0.5,0.5}
\definecolor{codepurple}{rgb}{0.58,0,0.82}
\definecolor{backcolour}{rgb}{0.95,0.95,0.92}
 
\lstdefinestyle{mystyle}{
    backgroundcolor=\color{backcolour},   
    commentstyle=\color{codegreen},
    keywordstyle=\color{magenta},
    numberstyle=\tiny\color{codegray},
    stringstyle=\color{codepurple},
    basicstyle=\ttfamily\footnotesize,
    breakatwhitespace=false,         
    breaklines=true,                 
    captionpos=b,                    
    keepspaces=true,                 
    numbers=left,                    
    numbersep=5pt,                  
    showspaces=false,                
    showstringspaces=false,
    showtabs=false,                  
    tabsize=2
}
 
\acrodef{SPHERE}{Sensor Platform for HEalthcare in a Residential Environment}
\acrodef{EPSRC}{Engineering and Physical Sciences Research Council}
\acrodef{IRC}{Interdisciplinary Research Collaboration}
\acrodef{PIS}{Participant Information Sheet}
\acrodef{IoT}{Internet of Things}
\acrodef{ML}{Machine Learning}
\acrodef{API}{Application Programming Interface}
\acrodef{DSL}{Domain Specific Language}

\newcommand{\hs}{{\sc HyperStream}}
\newcommand{\plate}[1]{{\bf #1} plate}
\newcommand{\ie}{{\em i.e.}}
\newcommand{\eg}{{\em e.g.}}

\newcommand{\etal}{{\em et.~al.}}

\newcommand{\mongo}{{\sc MongoDB}}
\newcommand{\nuc}{Intel\textregistered{} NUC}
\newcommand{\apache}{Apache HTTP Server\texttrademark{}}
\newcommand{\activemq}{Apache ActiveMQ\texttrademark{}}
\newcommand{\azure}{Microsoft\textregistered{} Azure Stream Analytics}
\newcommand{\spark}{Apache Spark\texttrademark{} Streaming}

\newcommand{\tom}{Tom Diethe}
\newcommand{\meelis}{Meelis Kull}
\newcommand{\niall}{Niall Twomey}
\newcommand{\kacper}{Kacper Sokol}
\newcommand{\hao}{Hao Song}
\newcommand{\emma}{Emma Tonkin}
\newcommand{\miquel}{Miquel Perell{\'o}-Nieto}
\newcommand{\peter}{Peter Flach}
\newcommand{\amazon}{Amazon, Cambridge, CB1 2GA, UK}
\newcommand{\isl}{Intelligent Systems Laboratory, University of Bristol, BS8 1UB, UK}
\newcommand{\tartu}{University of Tartu, 50090 Tartu, Estonia}

\newcommand{\names}{%
    (\tom,      \amazon,   tdiethe@amazon.com),
    (\meelis,   \tartu, meelis.kull@ut.ee),
    (\niall,    \isl,   niall.twomey@bristol.ac.uk),
    (\kacper,   \isl,   k.sokol@bristol.ac.uk),
    (\hao,      \isl,   hao.song@bristol.ac.uk),
    (\miquel,   \isl,   mp15688@bristol.ac.uk),
    (\emma,     \isl,   emma.tonkin@bristol.ac.uk),
    (\peter,    \isl,   peter.flach@bristol.ac.uk)
}

\StrCount{\names}{(}[\numnames]
\def\getname(#1,#2,#3){#1}
\def\getinst(#1,#2,#3){#2}
\def\getemail(#1,#2,#3){#3}

\jmlrheading{1}{2019}{1-48}{8/19}{10/00}{diethe19a}{%
    \foreach \n [count=\ni] in \names {%
        \ifnum \ni < \numnames%
            \expandafter\getname\n, %
        \else%
            and \expandafter\getname\n%
        \fi%
    }%
}

\ShortHeadings{\hs{}: a Workflow Engine for Streaming Data}{Diethe \etal}
\firstpageno{1}

\begin{document}
\title{\hs{}: a Workflow Engine for Streaming Data}

\author{%
    \foreach \n [count=\ni] in \names {%
    \name \expandafter\getname\n \email \expandafter\getemail\n \\
    \addr \expandafter\getinst\n%
    \ifnum \ni < \numnames%
        \AND%
    \fi%
    }
}

\editor{A. N. Other}

\maketitle

\begin{abstract}
This paper describes \hs{}, a large-scale, flexible and robust software package, written in the Python language, for processing streaming data with workflow creation capabilities. \hs{} overcomes the limitations of other computational engines and provides high-level interfaces to execute complex nesting, fusion, and prediction both in online and offline forms in streaming environments. \hs{} is a general purpose tool that is well-suited for the design, development, and deployment of \acl*{ML} algorithms and predictive models in a wide space of sequential predictive problems.

Source code, installation instructions, examples, and documentation can be found at: \url{https://github.com/IRC-SPHERE/HyperStream}.
\end{abstract}


\newcommand{\myurl}[1]{\footnote{\url{#1}}}

\section{Introduction}
Scientific workflow systems are designed to compose and execute a series of computational or data manipulation operations (workflow) \citep{deelman2009workflows}. Workflows simplify the process of sharing and reusing such operations, and enable scientists to track the provenance of execution results and the workflow creation steps. Generally workflow managers are designed to work in offline (batch) mode. Well known examples are Kepler\myurl{https://kepler-project.org/} and Taverna\myurl{https://taverna.incubator.apache.org/}.

In streaming data scenarios, common to most industry segments and big data use cases, dynamic data is generated on a continual basis. Stream processing solutions have been receiving increasing interest \citep{garofalakis2016data}, popular examples including \spark\myurl{http://spark.apache.org/streaming/} and \azure\myurl{https://azure.microsoft.com/en-us/services/stream-analytics/}. 

\hs{} harnesses the rich environment provided by the Python language to provide both stream processing and workflow engine capabilities, while maintaining an easy-to-use \ac{API}. 
This answers a growing need for scientific streaming data analysis in both academic and industrial data intensive research, as well as in fields outside of core computer science, such as healthcare and smart environments. \hs{} differs from other related toolboxes in various respects: i) it is distributed under the permissive MIT license, encouraging its use in both academic and commercial settings; ii) it depends only on a small set of requirements 
to ease deployment; iii) it focuses on streaming data sources unlike most workflow engines; is suitable for limited resource environments such as found in \ac{IoT} and Fog computing scenarios \citep{bonomi2012fog}; and iv) it allows both online and offline computational modes unlike most streaming solutions.

\section{Features}
This software has been designed from the outset to be domain-independent, in order to provide maximum value to the wider community. Source code, issue tracking, installation instructions, examples, and documentation can be found on GitHub\myurl{https://github.com/IRC-SPHERE/HyperStream}, 
as well as a discussion room\myurl{https://gitter.im/IRC-SPHERE-HyperStream}.
\hs{} is currently supported with Python 2.7 and 3.6 on *ix platforms (\eg{} linux, OS-X) and Microsoft Windows. For ease of installation,
\href{https://www.docker.com}{Docker} 
containers are provided. 
\hs{} also makes use of continuous integration using \href{https://travis-ci.org/IRC-SPHERE/HyperStream}{Travis-{CI}}.

The core requirements for \hs{} are summarised as follows:
%
    \begin{enumerate}[noitemsep,parsep=0pt,partopsep=0pt,topsep=0pt]
        \item the capability to create complex interlinked workflows
        \item a computational engine that is designed to be ``compute-on-request'' \label{req:request}
        \item to be capable of storing the history of computation \label{req:history}
        \item a plugin system for user extensions
        \item to be able to operate in online and offline mode \label{req:modes}
        \item to be lightweight and have minimal requirements \label{req:lightweight}
    \end{enumerate}
%
(\ref{req:request}) and (\ref{req:history}) reduce unnecessary repeated computation, and enable full data pipeline provenance. One of the main motivating factors for (\ref{req:lightweight}) was that computations should be capable of being performed on minimal hardware, such as found in \ac{IoT} settings (see \Cref{sec:sphere} below).

\section{Design}
\hs{} is written in Python, and uses \mongo{} for the back-end. This means that all system configuration and persistence is in \mongo{}, although this does not mean that \hs{} is limited to \mongo{} for stream storage (see below). The system consists of two layers: the stream layer and the workflow layer, as described below.

\subsection{Stream Layer}
\begin{figure}[h]
    \centering
    \includegraphics[width=0.6\linewidth]{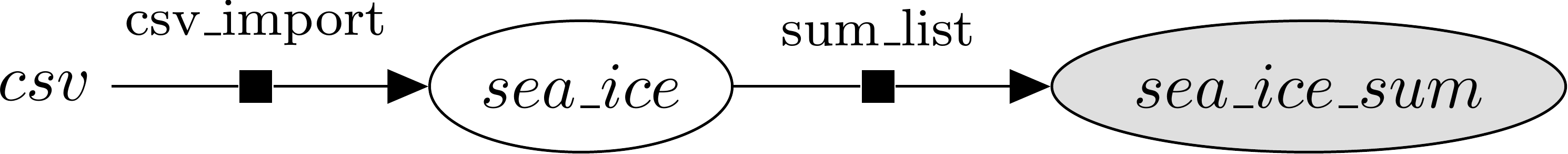}
    \caption{Example chain of computations. The filled (grey) node indicates that the $sea\_ice\_sum$ stream is stored in the database rather than memory.}
    \label{fig:example_chain}
\end{figure}

At the stream layer there exist only \emph{streams} and \emph{tools}. Tools operate on streams to produce new streams, hence creating a chain of operation. A simple example of this can be seen in \Cref{fig:example_chain}\myurl{https://github.com/IRC-SPHERE/HyperStream/blob/master/examples/tutorial_03.ipynb}. Here the data originates from a comma-separated value file, is imported using the $csv\_import$ tool into the $sea\_ice$ memory stream, and then processed by the $sum\_list$ tool into a database stream called $sea\_ice\_sum$. 

\vspace{-1em}

We treat all data as \emph{streams} of \emph{documents}, where a \emph{document} can contain most Python object types, as long as they can be converted to Binary JavaScript Object Notation (BSON). 
%

\emph{Tools} are the computation elements, with fixed parameters and filters defined that can reduce the amount of data that needs to be read from the database. They take input data in a standard format (an iterator over stream documents) and output data via a generator in the same standard format. Tools are agnostic to where the data actually lives (\ie{} memory/files/database). 

\emph{Channels} define the manifestation of streams for time ranges that have been computed, along with any specific processing required to read and write the streams, which abstracts away the specifics of interacting with different data sources. The built-in channels are the memory, database (\mongo{}), file, assets, (Python) module, and tool channels. The tool channel is a subclass of the module channel, which in turn is a subclass of the file channel, which means that the tools themselves are stored in streams.
The \hs{} plugin system allows users to define their own channels, in order to work with custom databases, file-based storage with custom formats or locations, or to modify the default capabilities of existing channels. An example machine learning plugin can be found at \myurl{https://github.com/IRC-SPHERE/HyperStreamOnlineLearning}. This wraps Scikit-learn \citep{scikit-learn} linear models into a HyperStream plugin, and provides examples for how this would be used for online learning and anomaly detection, and can for example be extended to the continual learning setting \citep{diethe2019continual}. 

\subsection{Workflow Layer}\label{sec:workflow}
Taking inspiration from factor graph notation for probabilistic graphical models \citep{buntine1994operations}, workflows define a graph of ``nodes'' connected by ``factors'', which can be surrounded by ``plates''. Workflows can have multiple time ranges, which will cause the streams contained in the nodes to be computed on all of the ranges given. Workflows can be defined to be operable in offline-only mode, or also available to the \hs{} online engine, which will cause the workflow to be executed continuously. Workflows are serialised to \mongo{} by \hs{} for ease of deployment.

\subsubsection{Plates}\label{sec:plates}
\textit{Plates} can be thought of as a ``for loop'' over parts of the computational graph contained within them. This is conceptually similar to the notion of plates in factor graphs. Both nodes 
and factors 
can be contained inside a plate.

An example is given below, where we construct an outer plate `C' that loops over continents, and then an inner plate that loops over countries within each continent:

\begin{lstlisting}[language=Python]
countries_dict = {
    'Asia': ['Bangkok', 'HongKong', 'KualaLumpur', 'NewDelhi', 'Tokyo'],
    'Australia': ['Brisbane', 'Canberra', 'GoldCoast', 'Melbourne',  'Sydney'],
    'NZ': ['Auckland', 'Christchurch', 'Dunedin', 'Hamilton','Wellington'],
    'USA': ['Chicago', 'Houston', 'LosAngeles', 'NY', 'Seattle']
}

# delete_plate requires the deletion to be first childs and then parents
for plate_id in ['C.C', 'C']:
    if plate_id in [plate[0] for plate in hs.plate_manager.plates.items()]:
        hs.plate_manager.delete_plate(plate_id=plate_id, delete_meta_data=True)

for country in countries_dict:
    id_country = 'country_' + country
    if not hs.plate_manager.meta_data_manager.contains(identifier=id_country):
        hs.plate_manager.meta_data_manager.insert(
            parent='root', data=country, tag='country', identifier=id_country)
    for city in countries_dict[country]:
        id_city = id_country + '.' + 'city_' + city
        if not hs.plate_manager.meta_data_manager.contains(identifier=id_city):
            hs.plate_manager.meta_data_manager.insert(
                parent=id_country, data=city, tag='city', identifier=id_city)
            
C = hs.plate_manager.create_plate(plate_id="C", description="Countries", values=[], complement=True,
                                  parent_plate=None, meta_data_id="country")
CC = hs.plate_manager.create_plate(plate_id="C.C", description="Cities", values=[], complement=True,
                                   parent_plate="C", meta_data_id="city")
\end{lstlisting}

\subsubsection{Nodes}\label{sec:nodes}
A \textit{node} is a collection of streams that live on the same plate, \ie{} they have shared meta-data keys, and are connected in the computational graph by factors.

\subsubsection{Factors}\label{sec:factors}
\textit{Factors} are the workflow implementations of tools. A factor defines the element of computation: the tool along with the source and sink nodes. Basic factors will take input streams on (a) given plate(s), execute the tool on these streams, and output a stream that is one same plate(s). Multi-output factors are able to take streams from a plate and output streams on a sub-plate of that plate (\eg{} by splitting). 

Usually, the first factor in a workflow will be a special ``raw'' factor that uses a tool with no input streams that pulls in data from a custom data source outside of \hs{}. 

\section{Domain Specific Languages}
\hs{} workkflows are defined in terms of a \ac{DSL}. 

\subsection{Case Study: The \acs{SPHERE} Project} \label{sec:sphere}
The \ac{SPHERE} project \citep{zhu2015bridging,woznowski2017sphere,Diethe:2018:REA:3219819.3219883} uses multiple heterogeneous sensors for the purpose of health monitoring within the home environment. Part of the project involves the deployment of sensor systems to the homes of up to 100 volunteer families within the Bristol area of the UK. Each house has a limited computational budget due to the inconvenience of off-site hardware installation. In this setting a \hs{} instance runs in each house in \emph{online} mode on an \nuc, a $4 \times 4$ inch mini PC, alongside other services such as \mongo, \activemq, and \apache. Here \hs{} is used to provide pseudo-real-time predictions using trained \acl{ML} models and to perform online processing and summarising of the sensor data. In addition, when data is retrieved from the houses, computations are performed on a centralised database to perform aggregate computations and further ``meta-summaries''. The \ac{SPHERE} project has made heavy use of the workflow capabilities and the plugin architecture of \hs{}.

\begin{figure}
    \centering
    \includegraphics[width=\linewidth]{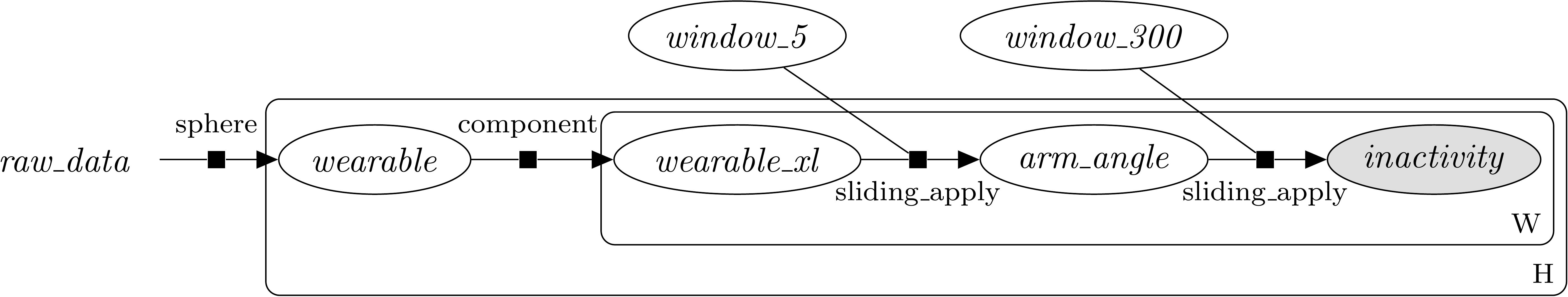}
    \caption{Example workflow.}
    \label{fig:example_workflow}
\end{figure}

\Cref{fig:example_workflow} 
depicts an example workflow for prediction of sleep, showing nested plates, nodes and factors. Here the raw data comes from the SPHERE deployment houses, which are on the \plate{H}. The wearable data is then split by its unique identifier (since there is more than one wearable per house) onto the \plate{W}, which is nested inside the \plate{H}. Two \texttt{sliding\_apply} tools are then executed for each wearable in each house with differing length sliding windows (5s and 300s) to first compute windowed arm angles and then a windowed inactivity estimate, which is stored in the database channel and subsequently used as part of a sleep prediction algorithm. 

\section{Concluding remarks}
We have presented \hs{}, a software package for processing streaming data with workflow creation capabilities, with a flexible plugin architecture. \hs{} is in active development, and contributions are actively welcomed (see \burl{https://github.com/IRC-SPHERE/HyperStream/wiki/How-to-contribute}). Moreover, we would like to acknowledge all \hs{} contributors, who can be identified using the \texttt{git log} command.

{The \ac{SPHERE} \acf{IRC} is funded by the UK \acf{EPSRC} under Grant EP/K031910/1.}
%

\small
\bibliography{diethe17a}

\end{document}